\documentclass{article} 
\usepackage{nips15submit_e,times}
\usepackage{cite}
\usepackage[cmex10]{amsmath}
\usepackage{fixltx2e}
\usepackage{datetime}
\usepackage{graphicx}
\usepackage{stfloats}
\usepackage{color}
\usepackage{amssymb}

\usepackage{pstricks,epsfig}
\usepackage{rotating}
\usepackage{subfig}
\usepackage{tikz}
\usepackage{enumerate}
\usepackage{tabularx}

\usepackage{pgf}
\usepackage{url}
\usepackage[margin=1.75cm]{geometry}
\usepackage{epstopdf}
\usepackage{color,soul}
\newcommand{\el}{{\it et al.}}

\title{StochasticNet: Forming Deep Neural Networks via Stochastic Connectivity}

\author{
Mohammad~Javad~Shafiee \\
Department of Systems Design Engineering\\
University of Waterloo\\
Ontario, Canada, N2L 3G1\\
\texttt{mjshafiee@uwaterloo.ca} \\
\And
Parthipan~Siva \\
Aimetis Corp.  \\
Waterloo, Ontario, Canada, N2L 4E9\\
\texttt{parthipan.siva@aimetis.com} \\
\And
Alexander~Wong \\
Department of Systems Design Engineering\\
University of Waterloo\\
Ontario, Canada, N2L 3G1\\
\texttt{a28wong@uwaterloo.ca}
}

\nipsfinalcopy 

\begin{document}

\maketitle

\begin{abstract}
Deep neural networks is a branch in machine learning that has seen a meteoric rise in popularity due to its powerful abilities to represent and model high-level abstractions in highly complex data.  One area in deep neural networks that is ripe for exploration is neural connectivity formation.  A pivotal study on the brain tissue of rats found that synaptic formation for specific functional connectivity in neocortical neural microcircuits can be surprisingly well modeled and predicted as a random formation.  Motivated by this intriguing finding, we introduce the concept of StochasticNet, where deep neural networks are formed via stochastic connectivity between neurons. As a result, any type of deep neural networks can be formed as a StochasticNet by allowing the neuron connectivity to be stochastic. Stochastic synaptic formations, in a deep neural network architecture, can allow for efficient utilization of neurons for performing specific tasks.  To evaluate the feasibility of such a deep neural network architecture, we train a StochasticNet using four different image datasets (CIFAR-10, MNIST, SVHN, and STL-10).  Experimental results show that a StochasticNet, using less than half the number of neural connections as a conventional deep neural network, achieves comparable accuracy and reduces overfitting on the CIFAR-10, MNIST and SVHN dataset. Interestingly, StochasticNet with less than half the number of neural connections, achieved a higher accuracy (relative improvement in test error rate of $\sim$6\% compared to ConvNet) on the STL-10 dataset than a conventional deep neural network. Finally, StochasticNets have faster operational speeds while achieving better or similar accuracy performances.
\end{abstract}

\section{Introduction}

Deep neural networks is a branch in machine learning that has seen a meteoric rise in popularity due to its powerful abilities to represent and model high-level abstractions in highly complex data.  Deep neural networks have shown considerable capabilities for handling specific complex tasks such as speech recognition~\cite{Hannun,Dahl}, object recognition~\cite{Krizhevsky,He,LeCun,Simonyan}, and natural language processing~\cite{Collobert,Bengio}.  Recent advances in improving the performance of deep neural networks have focused on areas such as network regularization~\cite{Zeller,Wan}, activation functions~\cite{Glorot1,Glorot2,He2}, and deeper architectures~\cite{Simonyan,Szegedy,Zhang}.  However, the neural connectivity formation of deep neural networks has remained largely the same over the past decade and thus further exploration and investigation on alternative approaches to neural connectivity formation can hold considerable promise.

To explore alternate deep neural network connectivity formation, we take inspiration from nature by looking at the way brain develops synaptic connectivity between neurons.  Recently, in a pivotal paper by Hill \el~\cite{Hill}, data of living brain tissue from Wistar rats was collected and used to construct a partial map of a rat brain.  Based on this map, Hill \el came to a very surprising conclusion. The synaptic formation, of specific functional connectivity in neocortical neural microcircuits, can be modelled and predicted as a random formation.  In comparison, for the construction of deep neural networks, the neural connectivity formation is largely deterministic and pre-defined.

Motivated by Hill \el's finding of random neural connectivity formation, we aim to investigate the feasibility and efficacy of devising stochastic neural connectivity formation to construct deep neural networks.  To achieve this goal, we introduce the concept of StochasticNet, where the key idea is to leverage random graph theory~\cite{Gilbert,Erdos} to form deep neural networks via stochastic connectivity between neurons.  As such, we treat the formed deep neural networks as particular realizations of a random graph.  Such stochastic synaptic formations in a deep neural network architecture can potentially allow for efficient utilization of neurons for performing specific tasks.  Furthermore, since the focus is on neural connectivity, the StochasticNet architecture can be used directly like a conventional deep neural network and benefit from all of the same approaches used for conventional networks such as data augmentation, stochastic pooling, and Dropout~\cite{dropout}, and DropConnect~\cite{dropconnect}.

While a number of stochastic strategies for improving deep neural network performance have been previously introduced~\cite{dropout,dropconnect,hashnet}, it is very important to note that the proposed StochasticNets is fundamentally different from these existing stochastic strategies in that StochasticNets' main significant contributions deals primarily with the formation of neural connectivity of individual neurons to construct efficient deep neural networks that are inherently sparse \textbf{prior} to training, while previous stochastic strategies deal with either the grouping of existing neural connections to explicitly enforce sparsity~\cite{hashnet}, or removal/introduction of neural connectivity for regularization \textbf{during} training.  More specifically, StochasticNets is a realization of a random graph formed prior to training and as such the connectivity in the network are \textbf{inherently sparse}, and are \textbf{permanent} and do not change during training.  This is very different from Dropout~\cite{dropout} and DropConnect~\cite{dropconnect} where the activations and connections are temporarily removed during training and put back during test for regularization purposes only, and as such the resulting neural connectivity of the network remains dense.  There is no notion of 'dropping' in StochasticNets as only a subset of possible neural connections are formed in the first place prior to training, and the resulting network connectivity of the network is sparse.

StochasticNets are also very different from HashNets~\cite{hashnet}, where connection weights are randomly grouped into hash buckets, with each bucket sharing the same weights, to explicitly sparsifying into the network, since there is no notion of grouping/merging in StochasticNets; the formed StochasticNets are naturally sparse due to the formation process.  In fact, stochastic strategies such as HashNets, Dropout, and DropConnect can be used \textbf{in conjunction} with StochasticNets.

The paper is organized as follows. First, a review of random graph theory is presented in Section 2.  The theory and design considerations behind forming StochasticNet as a random graph realizations are discussed in Section 3.  Experimental results using four image datasets (CIFAR-10~\cite{CIFAR10}, MNIST~\cite{MNIST}, SVHN~\cite{SVHN}, and STL-10~\cite{STL10}) to investigate the efficacy of StochasticNets with respect to different number of neural connections as well as different training set sizes is presented in Section 5.  Finally, conclusions are drawn in Section 6.

\section{Review of Random Graph Theory}

In this study, the goal is to leverage random graph theory~\cite{Gilbert,Erdos} to form the neural connectivity of deep neural networks in a stochastic manner.  As such, it is important to first provide a general overview of random graph theory for context.  In random graph theory, a random graph can be defined as the probability distribution over graphs~\cite{Bollobas}.  A number of different random graph models have been proposed in literature.

\begin{figure}
\vspace{- 0.3 cm}
\begin{center}
\includegraphics[scale = 0.2]{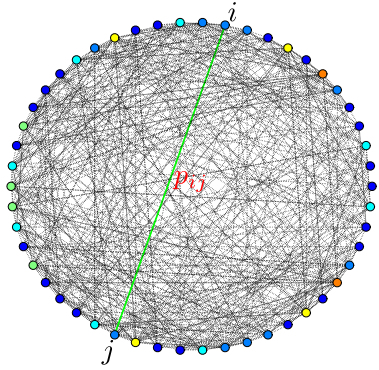}
\caption{An illustrative example of a random graph. All possible edge connectivity between the nodes in the graph may occur independently with a probability of $p_{ij}$.  }
\label{fig:RG}
\end{center}
\vspace{-0.4 cm}
\end{figure}

A commonly studied random graph model is that proposed by Gilbert~\cite{Gilbert}, in which a random graph can be expressed by $\mathcal{G}(n,p)$, where all possible edge connectivity are said to occur independently with a probability of $p$, where $0 < p < 1$.  This random graph model was generalized by Kovalenko~\cite{Kovalenko}, in which the random graph can be expressed by $\mathcal{G}(\mathcal{V},p_{ij})$, where $\mathcal{V}$ is a set of vertices and the edge connectivity between two vertices $\{i, j\}$ in the graph is said to occur with a probability of $p_{ij}$, where $0 < p_{ij} < 1$.  An illustrative example of a random graph based on this model is shown in Figure~\ref{fig:RG}. It can be seen that all possible edge connectivity between the nodes in the graph may occur independently with a probability of $p_{ij}$.

\begin{figure}
\vspace{- 0.2 cm}
\begin{center}
\begin{tabular}{cccc}
\includegraphics[scale = 0.13]{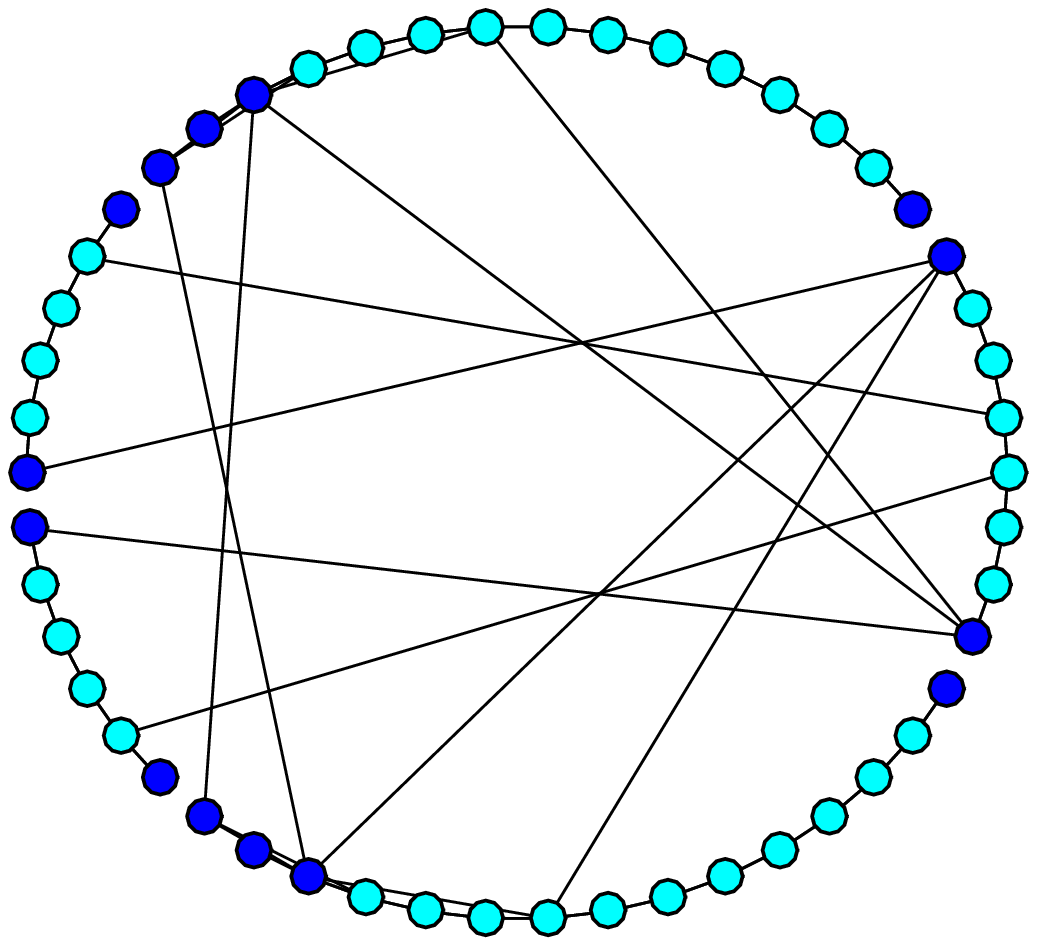}&
\includegraphics[scale = 0.13]{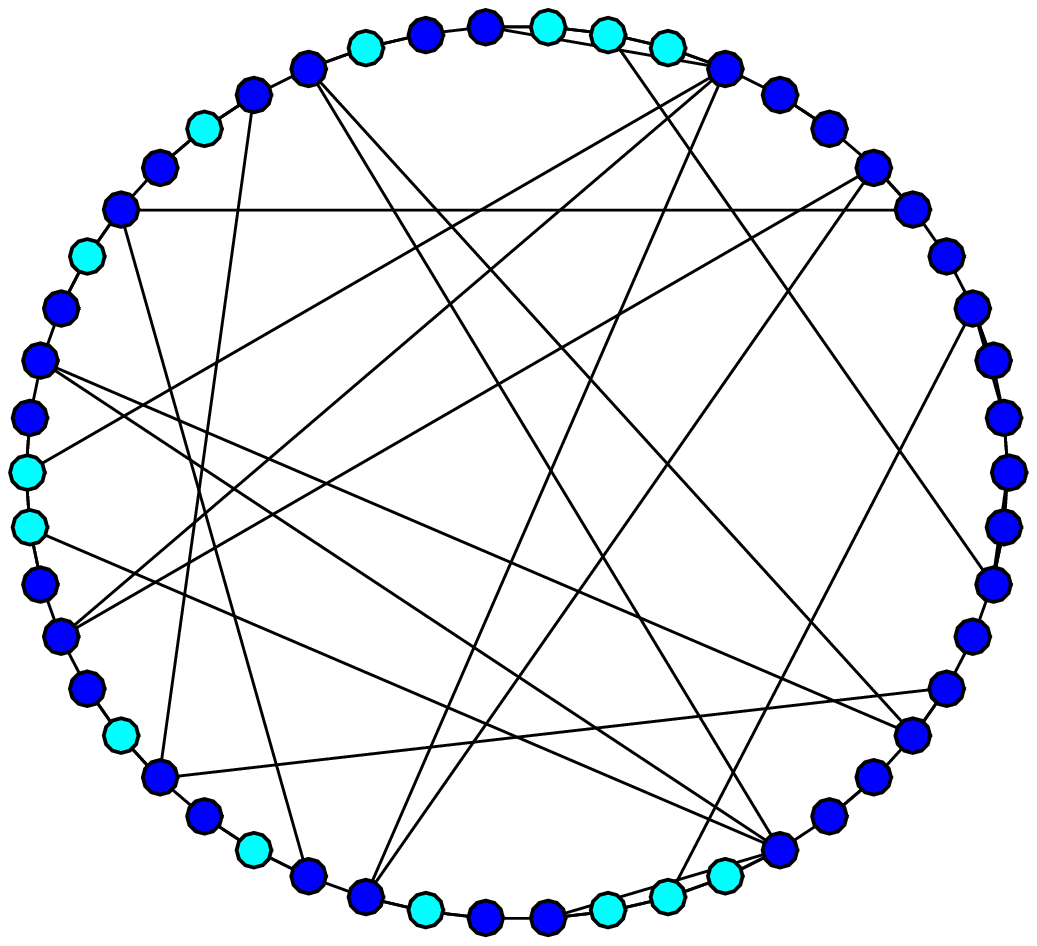}&
\includegraphics[scale = 0.13]{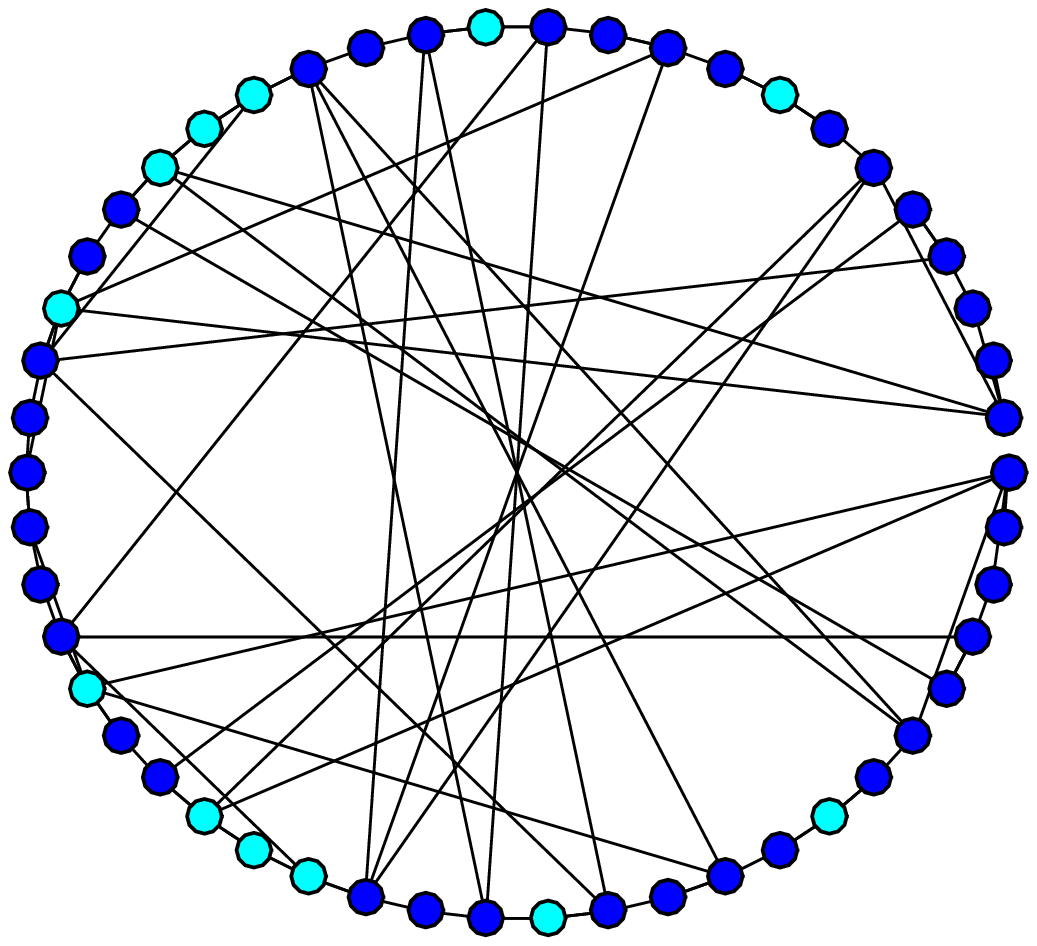}&
\includegraphics[scale = 0.13]{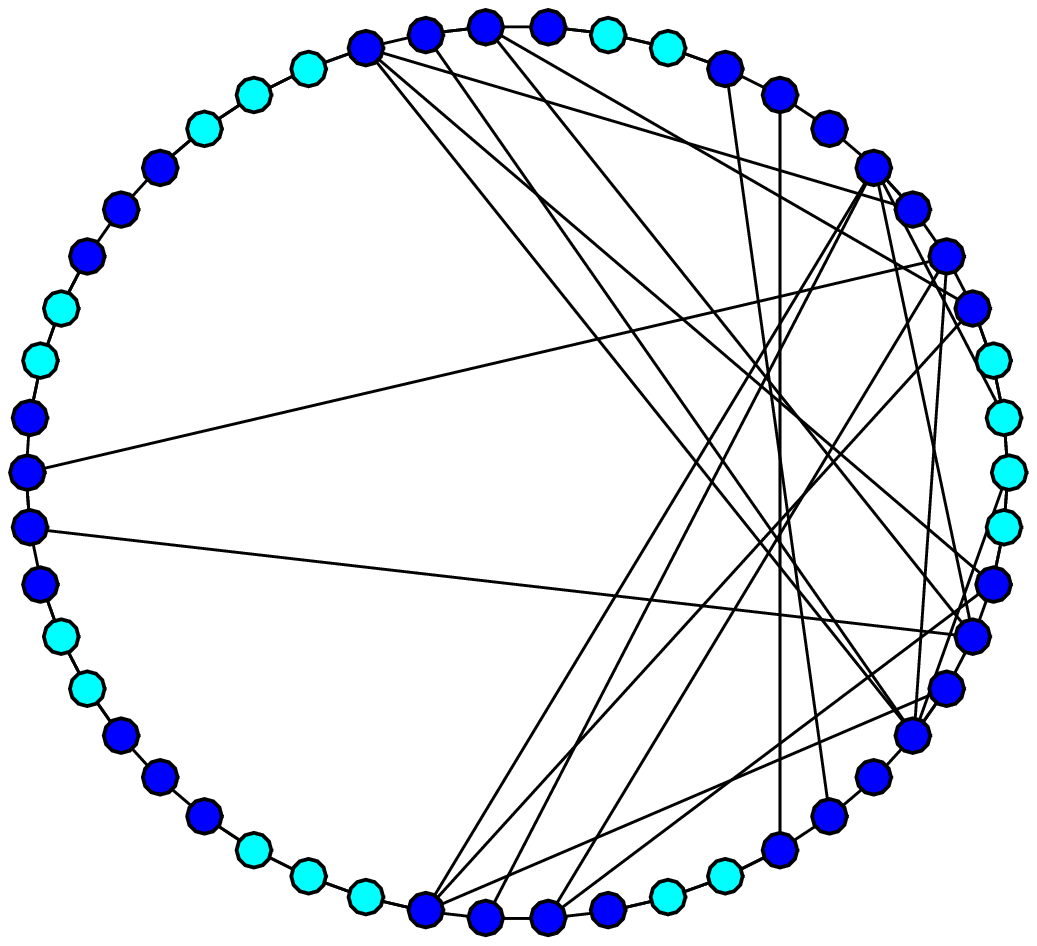}
\end{tabular}
\vspace{-0.1cm}
\caption{ Realizations of random graph in Figure~\ref{fig:RG}. The probability for edge connectivity between all nodes in the graph was set to $p_{i,j}=0.1$ for all nodes $i$ and $j$. Each diagram demonstrates a different realization of the random graph. }
\label{fig:RGrealization}
\end{center}
\vspace{-0.4 cm}
\end{figure}

Therefore, based on this generalized random graph model, realizations of random graphs can be obtained by starting with a set of $n$ vertices $\mathcal{V} = \{v_q|1 \geq q \geq n\}$ and randomly adding a set of edges between the vertices based on the set of possible edges $\mathcal{E} = \{e_{ij}|1 \geq i \geq n, 1 \geq j \geq n\}$ independently with a probability of $p_{ij}$.  A number of realizations of the random graph in Figure~\ref{fig:RG} are provided in Figure~\ref{fig:RGrealization} for illustrative purposes. It is worth noting that because of the underlying probability distribution, the generated realizations of the random graph often exhibit differing edge connectivity.

Given that deep neural networks can be fundamentally expressed and represented as graphs $\mathcal{G}$, where the neurons are vertices $\mathcal{V}$ and the neural connections are edges $\mathcal{E}$, one intriguing idea for introducing stochastic connectivity for the formation of deep neural networks is to treat the formation of deep neural networks as particular realizations of random graphs, which we will describe in greater detail in the next section.

\section{StochasticNets: Deep Neural Networks as Random Graph Realizations}

Let us represent the full network architecture of a deep neural network as a random graph $\mathcal{G}(\mathcal{V},p[^{i \rightarrow j}_{k\rightarrow h}])$, where $\mathcal{V}$ is the the set of neurons $\mathcal{V} = \{v_{i,k}|1 \geq i \geq n_{l}, 1 \geq k \geq m_{i}\}$, with $v_{i,k}$ denoting the $k^{\rm th}$ neuron at layer $i$, $n_{l}$ denoting the number of layers, $m_{i}$ denoting the number of neurons at layer $i$, and $p[^{i \rightarrow j}_{k\rightarrow h}]$ is the probability that a neural connection occurs between neuron $v_{j,h}$ and $v_{i,k}$.

Based on the above random graph model for representing deep neural networks, one can then form a deep neural network as a realization of the random graph $\mathcal{G}(\mathcal{V},p[^{i \rightarrow j}_{k\rightarrow h}])$ by starting with a set of neurons $\mathcal{V}$, and randomly adding neural connections between the set of neurons independently with a probability of $p[^{i \rightarrow j}_{k\rightarrow h}]$ as defined above.

\begin{figure}
\vspace{- 0.3 cm}
\begin{center}
\includegraphics[scale = 0.12]{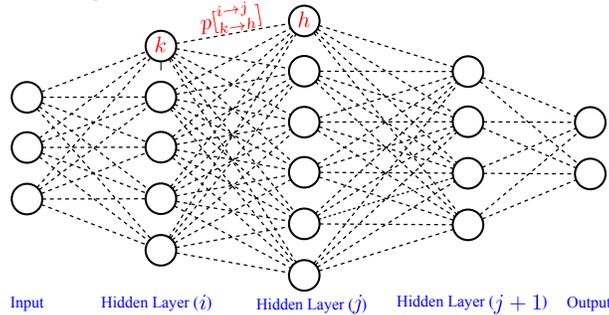}
\caption{Example random graph representing a general deep feed-forward neural network. Every neuron $k$ in layer $i$ may be connected to neuron $h$ in layer $j$ with probability $p[^{i \rightarrow j}_{k\rightarrow h}]$ based on random graph theory.  To enforce the properties of a general deep feed-forward neural network, $p[^{i \rightarrow j}_{k\rightarrow h}] = 0$ when $i=j~||~|i-j|>2$.}
\label{fig:StochasticNetRG}
\end{center}
\vspace{-0.4 cm}
\end{figure}

\begin{figure}[t]
\vspace{- 0.3 cm}
\begin{center}
\includegraphics[scale = 0.12]{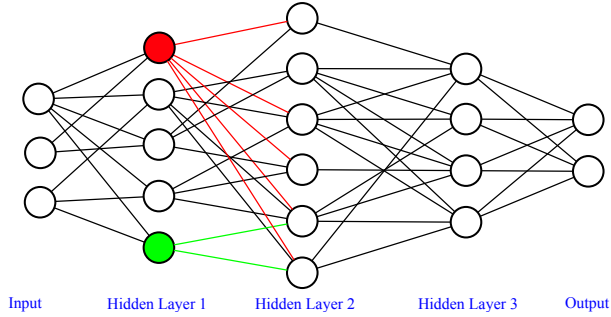}
\caption{ An example realization of the random graph shown in Figure~\ref{fig:StochasticNetRG}. In this example, $p[^{i \rightarrow j}_{k\rightarrow h}] = 0.5$ for all neurons except when $i=j~||~|i-j|>2$.  It can be observed that the neural connectivity for each neuron may be different due to the stochastic nature of neural connection formation. The connectivity for the red neuron and the green neuron are highlighted to show the differences in neural connectivity.}
\label{fig:StochasticNet}
\end{center}
\vspace{-0.4 cm}
\end{figure}

While one can form practically any type of deep neural network as a random graph realizations, an important design consideration for forming deep neural networks as random graph realizations is that different types of deep neural networks have fundamental properties in their network architecture that must be taken into account and preserved in the random graph realization.  Therefore, to ensure that fundamental properties of the network architecture of a certain type of deep neural network is preserved, the probability $p[^{i \rightarrow j}_{k\rightarrow h}]$ must be designed in such a way that these properties are enforced appropriately in the resultant random graph realization.  Let us consider a general deep feed-forward neural network.  First, in a deep feed-forward neural network, there can be no neural connections between non-adjacent layers.  Second, in a deep feed-forward neural network, there can be no neural connections between neurons on the same layer.  Therefore, to enforce these two properties, $p[^{i \rightarrow j}_{k\rightarrow h}] = 0$ when $i=j~||~|i-j|>2$.  An example random graph based on this random graph model for representing general deep feed-forward neural networks is shown in Figure~\ref{fig:StochasticNetRG}, with an example realization of the random graph shown in Figure~\ref{fig:StochasticNet}.  It can be observed in Figure~\ref{fig:StochasticNet} that the neural connectivity for each neuron may be different due to the stochastic nature of neural connection formation.

Furthermore, for specific types of deep feed-forward neural networks, additional considerations must be taken into account to preserve their properties in the resultant random graph realization.  For example, in the case of deep convolutional neural networks, neural connectivity in the convolutional layers are arranged such that small spatially localized neural collections are connected to the same output neuron in the next layer.  Furthermore, the weights of the neural connections are shared amongst different small neural collections.  A significant benefit to this architecture is that it allows neural connectivity at the convolutional layers to be efficiently represented by a set of local receptive fields, thus greatly reducing memory requirements and computational complexity.  To enforce these properties when forming deep convolutional neural networks as random graph realizations, one can further enforce the probability $p[^{i \rightarrow j}_{k\rightarrow h}]$ such that the probability of neural connectivity is defined at a local receptive field level.  As such, the neural connectivity for each randomly realized local receptive field is based on a probability distribution, with the neural connectivity configuration thus being shared amongst different small neural collections for a given randomly realized local receptive field.

\begin{figure}
\vspace{- 0.3 cm}
\begin{center}
\includegraphics[scale = 0.1]{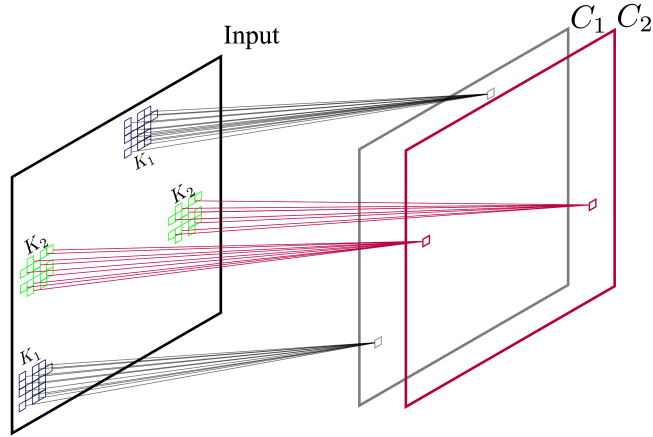}
\caption{ Forming a deep convolutional neural network from a random graph.  The neural connectivity for each randomly realized local receptive field $\{K_{1},K_{2}\}$ are determined based on a probability distribution, and as such the configuration and shape of each randomly realized local receptive field may differ.  It can be seen that the shape and neural connectivity for local receptive field $K_{1}$ is completely different from local receptive field $K_{2}$.  The response of each randomly realized local receptive field leads to an output in new channel $C$. Only one layer of the formed deep convolutional neural network from a random graph is shown for illustrative purposes.   }
\label{fig:StochasticCNN}
\end{center}
\vspace{-0.4 cm}
\end{figure}

Given this random graph model for representing deep convolutional neural networks, the resulting random graph realization is a deep convolutional neural network where each convolutional layer consists of a set of randomly realized local receptive fields $K$, with each randomly realized local receptive field $K_{i,k}$, which denotes the $k^{\rm th}$ receptive field at layer $i$, consisting of neural connection weights of a set of random neurons within a small neural collection to the output neuron.  An example of a realization of a deep convolutional neural network from a random graph is shown in Figure~\ref{fig:StochasticCNN}.

\section{Experimental Results}

\subsection{Experimental Setup}

To investigate the efficacy of StochasticNets, we construct StochasticNets with a deep convolutional neural network architecture and evaluate the constructed StochasticNets in a number of different ways.  First, we investigate the effect of the number of neural connections formed in the constructed StochasticNets on its performance for the task of image object recognition.  Second, we investigate the performance of StochasticNets when compared to baseline deep convolutional neural networks (which we will simply refer to as ConvNets) with standard neural connectivity for different image object recognition tasks based on different image datasets.  Third, we investigate the relative speed of StochasticNets during classification with respect to the number of neural connections formed in the constructed StochasticNets.  It is important to note that the main goal is to investigate the efficacy of forming deep neural networks via stochastic connectivity in the form of StochasticNets and the influence of stochastic connectivity parameters on network performance, and not to obtain maximum absolute performance; therefore, the performance of StochasticNets can be further optimized through additional techniques such as data augmentation and network regularization methods.  For evaluation purposes, four benchmark image datasets are used: CIFAR-10~\cite{CIFAR10}, MNIST~\cite{MNIST}, SVHN~\cite{SVHN}, and STL-10~\cite{STL10}.  A description of each dataset and the StochasticNet configuration used are described below.

\subsubsection{Datasets}

The CIFAR-10 image dataset~\cite{CIFAR10} consists of 50,000 training images categorized into 10 different classes (5,000 images per class) of natural scenes. Each image is an RGB image that is 32$\times$32 in size.  The MNIST image dataset~\cite{MNIST} consists of 60,000 training images and 10,000 test images of handwritten digits.  Each image is a binary image that is 28$\times$28 in size, with the handwritten digits are normalized with respect to size and centered in each image.  The SVHN image dataset~\cite{SVHN} consists of 604,388 training images and 26,032 test images of digits in natural scenes. Each image is an RGB image that is 32$\times$32 in size.  The images in the MNIST dataset were resized to $32 \times 32$ by zero padding since the same StochasticNet network configuration is utilized for all mentioned image datasets.  Finally, the STL-10 image dataset~\cite{STL10} consists of 5,000 labeled training images and 8,000 labeled test images categorized into 10 different classes (500 training images and 800 training images per class) of natural scenes. Each image is an RGB image that is 96$\times$96 in size.  Note that the 100,000 unlabeled images in the STL-10 image dataset were not used in this study.

\subsubsection{StochasticNet Configuration}
\label{SNETCONFIG}
The StochasticNets used in this study for the all datasets are realized based on the LeNet-5 deep convolutional neural network~\cite{MNIST} architecture, and consists of 3 convolutional layers with 32, 32, and 64 local receptive fields of size $5 \times 5$ for the first, second, and third convolutional layers, respectively, and 1 hidden layer of 64 neurons, with all neural connections in the convolutional and hidden layers being randomly realized based on probability distributions.  While it is possible to take advantage of any arbitrary distribution to construct StochasticNet realizations, for the purpose of this study the neural connection probability of the hidden layers follow a uniform distribution, while two different spatial distributions were explored for the convolutional layers: i) uniform distribution, and ii) a Gaussian distribution with the mean at the center of the receptive field and the standard deviation being a third of the receptive field size.  All image datasets are with 10 class label outputs which is provided in the network setup.

\subsection{Number of Neural Connections}
\label{numconnections}
\begin{figure}
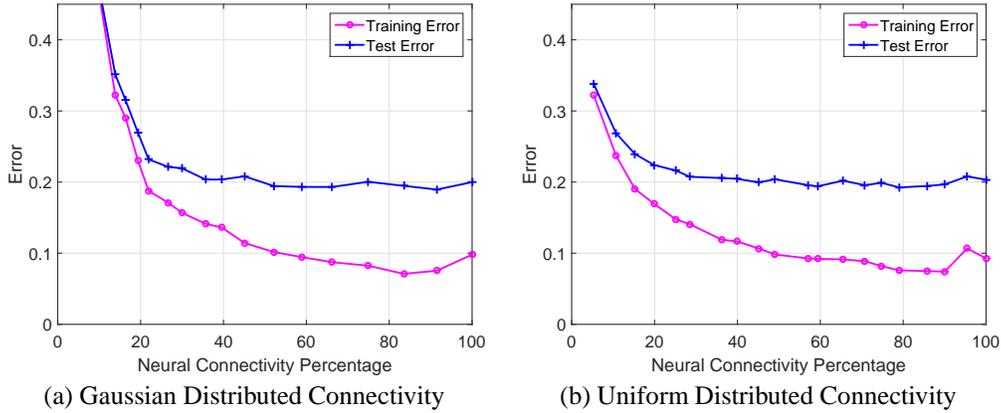

\vspace{- 0.3 cm}
\begin{center}
\begin{tabular}{cc}
\includegraphics[scale = 0.35]{TrainingVSCon_Gauss}&
\includegraphics[scale = 0.35]{TrainingVSCon_Uniform}\\
(a) Gaussian Distributed Connectivity & (b) Uniform Distributed Connectivity
\end{tabular}
\caption{Training and test error versus the number of neural connections in convolutional layers and fully connected layers for the CIFAR-10 dataset. Both Gaussian distributed and uniform distributed neural connectivity were evaluated. Note that neural connectivity percentage of 100 is equivalent to ConvNet, since all connections are made.}
\label{fig:TrainingVSCon}
\end{center}
\vspace{-0.4 cm}
\end{figure}

An experiment was conducted to illustrate the impact of the number of neural connections on the modeling accuracy of StochasticNets. Figure~\ref{fig:TrainingVSCon} demonstrates the training and test error versus the number of neural connections in the network for the CIFAR-10 dataset.  A StochasticNet with the network configuration as described in Section~\ref{SNETCONFIG} was provided to train the model. The neural connection probability is varied in both the convolutional layers and the hidden layer to achieve the desired number of neural connections for testing its effect on modeling accuracy.

Figure~\ref{fig:TrainingVSCon} demonstrates the training and testing error vs. the neural connectivity percentage relative to the baseline ConvNet, for two different neural connection distributions: i) uniform distribution, and ii) a Gaussian distribution with the mean at the center of the receptive field and the standard deviation being a third of the receptive field size.  It can be observed that StochasticNet is able to achieve the same test error as ConvNet when the number of neural connections in the StochasticNet is less than half that of the ConvNet.  It can be also observed that, although increasing the number of neural connections resulted in lower training error, it does not not exhibit reductions in test error, which brings to light the issue of over-fitting.  In other words, it can be observed that the proposed StochasticNets can improve the handling of over-fitting associated with deep neural networks while decreasing the number of neural connections, which in effect greatly reduces the number of computations and thus resulting in faster network training and usage.  Finally, it is also observed that there is a noticeable difference in the training and test errors when using Gaussian distributed connectivity when compared to uniform distributed connectivity, which indicates that the choice of neural connectivity probability distributions can have a noticeable impact on model accuracy.

\begin{figure}[!ht]
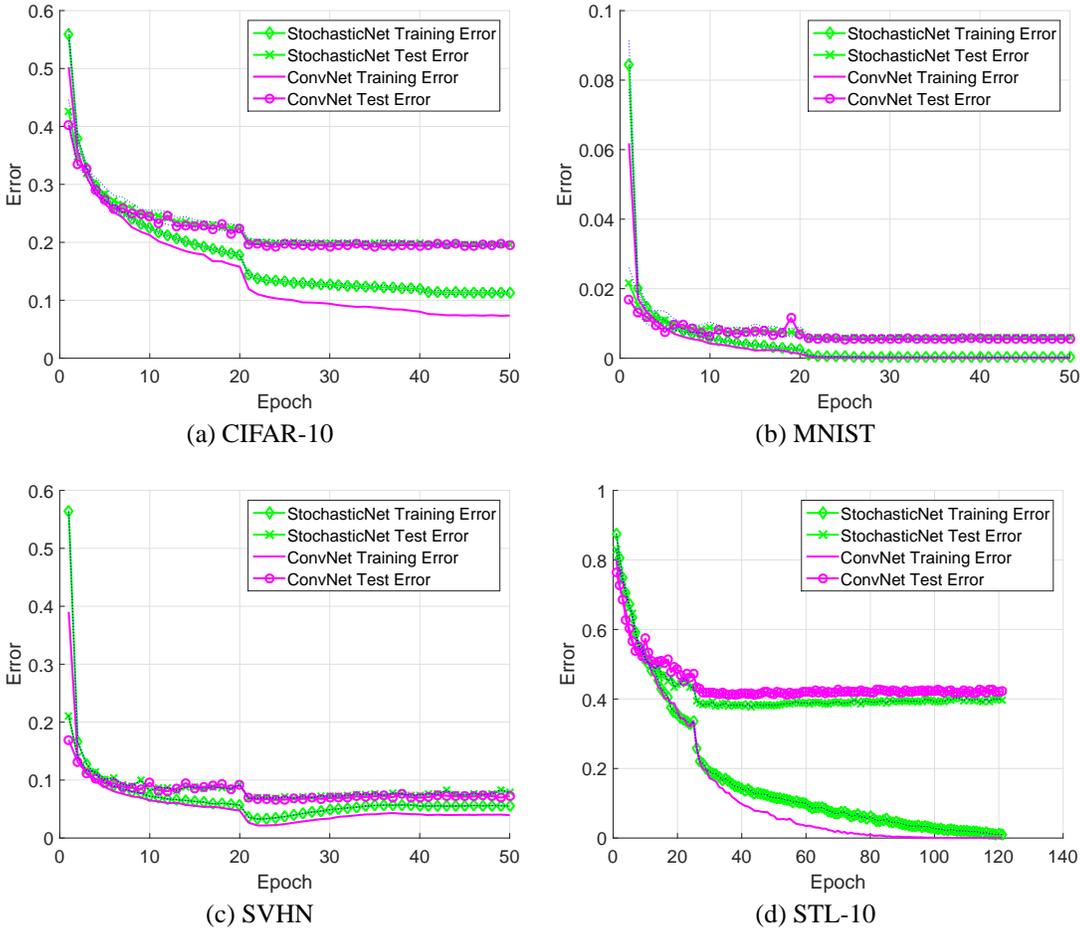

\begin{center}
\begin{tabular}{cc}
\includegraphics[scale = 0.38]{TrainingTrials_CIFAR}&
\includegraphics[scale = 0.38]{TrainingTrials_MNIST}\\
(a) CIFAR-10 & (b) MNIST\\
~ & ~\\
\includegraphics[scale = 0.38]{TrainingTrials_SVHN}&
\includegraphics[scale = 0.38]{TrainingTrials_STL10}\\
(c) SVHN & (d) STL-10\\
\end{tabular}
\caption{ Comparison between a standard ConvNet and a StochasticNet with 39\% of neural connectivity as the ConvNet. For StochasticNets, the results shows the error based on 25 trials since the neural connectivity of StochasticNets are realized stochastically. The dashed line demonstrates the standard deviation of error based on 25 trials for StochasticNets.   }
\label{fig:StochasticNetVSConvNet}
\end{center}
\end{figure}

\subsection{Comparisons with ConvNet}

Motivated by the results shown in Figure~\ref{fig:TrainingVSCon}, a comprehensive experiment were done to demonstrate the performance  of the proposed StochasticNets on different benchmark image datasets. StochasticNet realizations were formed with 39\% neural connectivity via Gaussian-distributed connectivity when compared to a conventional ConvNet. The StochasticNets and ConvNets were trained on four benchmark image datasets (i.e., CIFAR-10, MNIST, SVHN, and STL-10) and their training and test error performances are compared to each other. Since the neural connectivity of StochasticNets are realized stochastically, the performance of the StochasticNets was evaluated based on 25 trials (leading to 25 StochasticNet realizations) and the reported results are based on the average of the 25 trials. Figure~\ref{fig:StochasticNetVSConvNet} shows the training and test error results of the StochasticNets and ConvNets on the four different tested datasets.  It can be observed that, despite the fact that there are less than half as many neural connections in the StochasticNet realizations, the test errors between ConvNets and the StochasticNet realizations can be considered to be the same for CIFAR-10, MNIST, and SVHN datasets.  Interestingly, it was also observed that the test errors for the StochasticNet realizations is lower than that achieved using the ConvNet (relative improvement in test error rate of $\sim$6\% compared to ConvNet) for the STL-10 dataset, again despite the fact that there are less than half as many neural connections in the StochasticNet realizations.  The results for the STL-10 dataset truly illustrates the particular effectiveness of StochasticNets, particularly when dealing with low number of training samples.

Furthermore, the gap between the training and test errors of the StochasticNets is less than that of the ConvNets, which would indicate reduced overfitting in the StochasticNets.  The standard deviation of the 25 trials for each error curve is shown as dashed lines around the error curve.  It can be observed that the standard deviation of the 25 trials is very small and indicates that the proposed StochasticNet exhibited similar performance in all 25 trials.

\subsection{Relative Speed vs. Number of Neural Connections}

\begin{figure}
\vspace{- 0.3 cm}
\begin{center}
\begin{tabular}{cc}
\includegraphics[scale = 0.35]{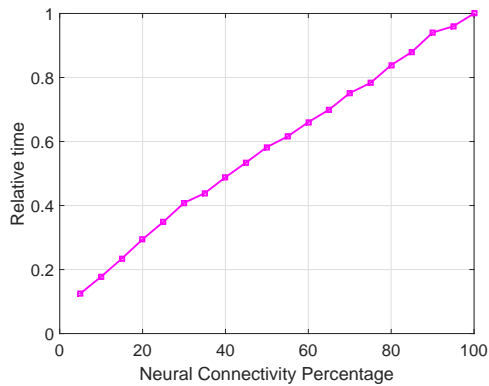}
\end{tabular}
\caption{Relative classification time versus the number of neural connections. Note that neural connectivity percentage of 100 is equivalent to ConvNet, since all connections are made.}
\label{fig:SpeedVSCon}
\end{center}
\vspace{-0.4 cm}
\end{figure}

Given that the experiments in the previous sections show that StochasticNets can achieve good performance relative to conventional ConvNets while having significantly fewer neural connections, we now further investigate the relative speed of StochasticNets during classification with respect to the number of neural connections formed in the constructed StochasticNets.  Here, as with Section~\ref{numconnections}, the neural connection probability is varied in both the convolutional layers and the hidden layer to achieve the desired number of neural connections for testing its effect on the classification speed of the formed StochasticNets.  Figure~\ref{fig:SpeedVSCon} demonstrates the relative classification time vs. the neural connectivity percentage relative to the baseline ConvNet.  The relative time is defined as the time required during the classification process relative to that of the ConvNet.  It can be observed that the relative time decreases as the number of neural connections decrease, which illustrates the potential for StochasticNets to enable more efficient classification.

%
%
%

%
%
%

\section{Conclusions}

In this study, we introduced a new approach to deep neural network formation inspired by the stochastic connectivity exhibited in synaptic connectivity between neurons.  The proposed StochasticNet is a deep neural network that is formed as a realization of a random graph, where the synaptic connectivity between neurons are formed stochastically based on a probability distribution.  Using this approach, the neural connectivity within the deep neural network can be formed in a way that facilitates efficient neural utilization, resulting in deep neural networks with much fewer neural connections while achieving the same modeling accuracy. The effectiveness and efficiency of the proposed StochasticNet was evaluated using four popular benchmark image datasets and compared to a conventional convolutional neural network (ConvNet). Experimental results demonstrate that the proposed StochasticNet provides comparable accuracy as the conventional ConvNet with much less number of neural connections while reducing the overfitting issue associating with the conventional ConvNet for CIFAR-10, MNIST, and SVHN datasets.  More interestingly, a StochasticNet with much less number of neural connections was found to achieve higher accuracy when compared to conventional deep neural networks for the STL-10 dataset.  As such, the proposed StochasticNet holds great potential for enabling the formation of much more efficient deep neural networks that have fast operational speeds while still achieving strong accuracy.

\section*{Acknowledgments}

This work was supported by the Natural Sciences and Engineering Research Council of Canada, Canada Research Chairs Program, and the Ontario Ministry of Research and Innovation. The authors also thank Nvidia for the GPU hardware used in this study through the Nvidia Hardware Grant Program.

\section*{Author contributions}

M.S. and A.W. conceived and designed the architecture.  M.S., P.S., and A.W. worked on formulation and derivation of architecture.  M.S. implemented the architecture and performed the experiments.  M.S., P.S., and A.W. performed the data analysis. All authors contributed to writing the paper and to the editing of the paper.

\end{document}